\documentclass{article}

\usepackage{arxiv}
\usepackage[utf8]{inputenc} 
\usepackage[T1]{fontenc}    
\usepackage{hyperref}       
\usepackage{url}            
\usepackage{booktabs}       
\usepackage{amsfonts}       
\usepackage{nicefrac}       
\usepackage{microtype}      
\usepackage{lipsum}
\usepackage{graphicx}
\usepackage{float}
\usepackage{amsfonts}
\usepackage{booktabs}
\usepackage{siunitx}
\usepackage{graphicx}
\usepackage[center]{caption}
\usepackage{subcaption}
\usepackage{array, makecell}
\usepackage{amsmath}

\title{Recurrent Neural Network based Electricity Load Forecasting of G-20 Members}

\author{
 Jaymin Suhagiya \\
  Department of Information and Communication Technology\\
  Adani Institute of Infrastructure Engineering\\
  Ahmedabad, Gujarat, India - 382421 \\
  \texttt{jayminsuhagiya.ict17@gmail.com} \\
   \And
   Deep Raval \\
  Department of Information and Communication Technology\\
  Adani Institute of Infrastructure Engineering\\
  Ahmedabad, Gujarat, India - 382421 \\
  \texttt{deepraval.ict17@gmail.com} \\
   \And
 Siddhi Vinayak Pandey \\
  Department of Electrical Engineering\\
  Adani Institute of Infrastructure Engineering\\
  Ahmedabad, Gujarat, India - 382421 \\
  \texttt{siddhipandey.ele17@aii.ac.in} \\
  \And
  Jeet Patel \\
  Department of Electrical Engineering\\
  Adani Institute of Infrastructure Engineering\\
  Ahmedabad, Gujarat, India - 382421 \\
  \texttt{jeetpatel.ele17@aii.ac.in} \\
  \And
 Ayushi Gupta \\
  Department of Electronics and Communication Engineering\\
 Pranveer Singh Institute of Technology \\ Kanpur, Uttar Pradesh, India - 209305\\
 ayushiaggra2001@gmail.com \\
 \And
 Akshay Srivastava \\
 Department of Electronics and Communication Engineering\\
 Pranveer Singh Institute of Technology\\ Kanpur, Uttar Pradesh, India - 209305\\
 mechengineerakshay@gmail.com \\
  
}

\begin{document}
\maketitle
\begin{abstract}
Forecasting the actual amount of electricity with respect to the need/demand of the load is always been a challenging task for each power plants based generating stations. Due to uncertain demand of electricity at receiving end of station causes several challenges such as: reduction in performance parameters of generating and receiving end stations, minimization in revenue, increases the jeopardize for the utility to predict the future energy need for a company etc. With this issues, the precise forecasting of load at the receiving end station is very consequential parameter to establish the impeccable balance between supply and demand chain. In this paper, the load forecasting of G-20 members have been performed utilizing the Recurrent Neural Network coupled with sliding window approach for data generation. During the experimentation we have achieved Mean Absolute Test Error of 16.2193 TWh using LSTM.
\end{abstract}


\section{Introduction}
\par An electrical load is a component of a circuit that consumes power, in other words the current drawn from the source by the components connected in parallel is called as an electrical load. The standard economic growth, population variance, geographical variations etc. of a country is additionally defined by the load consumption parameters. All the factors can be surmised and analyzed by load distribution and variation in a particular area. Demand forecast are habituated to determine the capacity of generation, transmission and distribution system. Generally, load connected in a circuit is not constant. If a person turns on a light bulb in a house there will be a slight increase in the input power of a generator in a power plant. Due to this there is always an uncertainty in the load. Consequentially, load forecasting is a very important part of a power system analysis.

\subsection{Why load forecasting is required}

\par As the use of electricity is increasing day by day; the load forecasting becomes important due to many reasons. One of the major reasons is that - nonrenewable sources of energy; which is decreasing drastically. While, the efficiency of renewable energy source; which is not a quite reliable in nature. Load forecasting also helps during the power system expansion which starts from the future load anticipations. If future increase of the load is needed, then cost and capacity of new power plant can be estimated. Load forecasting can also be used for safety purpose. In Industrial sector the load consumed is peak load most of the time, but there is always a limit for a particular industry above which they cannot draw the power from the grid or else they are charged very heavy for the carelessness .Load forecasting helps notify that the limit is going to be breached sometime soon.
 
\subsection{History \& Advancement in load forecasting:}

\par In 1800, electricity was discovered by an Italian Physicist, Alessandro Volta. This discovery boosted up the economy of world at the crowning point. Later on, researchers worked on electricity power-based gadgets. Subsequently, industries became dependent on electricity as it is the primary product of the power industry. People started discovering new methods for generation of electricity. But for estimation of electricity required by power industries any country, people claimed a method so that they can predict that how much electricity is required by power industries. These industries move the electrons through the grid to end users. There are two features that make the electric supply chain very different from other supply chains. Firstly, the electrons travel really fast at the speed of light and secondly, there is not yet a practical solution for bulk storage of electricity. Therefore, the production and consumption have to be balanced in real time. To operate and plan the system, people have to understand when, where and how much electricity is spread throughout the system. That is why load forecasting is crucial to the power industry \cite{hong2015load}. In 1880’s, the power companies simply used Layman method for predicting the use of electricity in future. They predicted the future consumption by simply developing an engineering approach to manually forecast the future load using charts and tables. Some of those elements, such as heating/cooling degree days, temperature-humidity index, and wind-chill factor, are inherited by today's load forecasting models. The similar day method, which derives a future load profile using the historical days with similar temperature profiles and day type (e.g., day of the week and holiday), is still used by many utilities \cite{ref4, ref5}.
    
\par In $1940s$, when air conditioners invented, the demand of electricity got tremendously affected by whether change and climate change. In winters, electricity usage got down but in summers it raised. After the discovery of air conditioners there came electric heaters for winters. So, prediction and estimation of electricity usage became very hectic by using the traditional method. The below mentioned graph shows how weather is affecting the electricity consumed by the load and this graph is plotted using advanced methods of load forecasting\cite{staffell2018increasing}. 

\begin{figure} [h!]
\hfil \includegraphics[width=8cm,height=7cm]{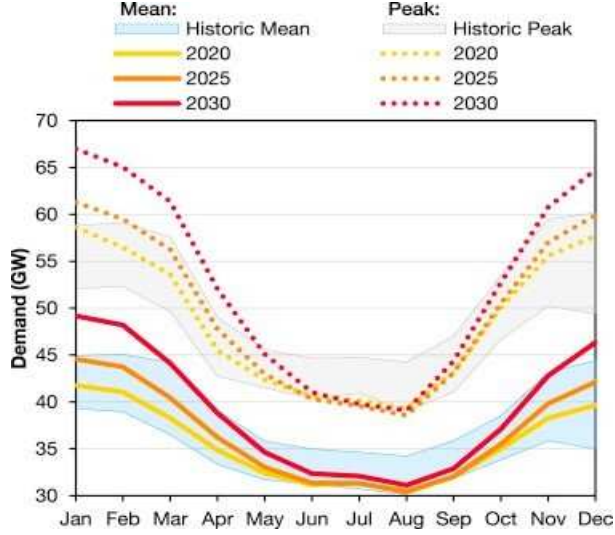}
\caption{The increasing impact on weather on electricity supply and demand \cite{staffell2018increasing}}\label{fig:cm}
\end{figure}

\par Presently, in this period of booming increment in innovation, demand of power has come to at apex \cite{ref8, alagbe2019artificial}, so industrialists require a few progressed strategies for load forecasting. And there are created strategies like short term load forecasting \cite{ref10} and medium-and long-term forecasting \cite{xiao2016combined}. The long-term load forecasting covers horizons of one to ten years \cite{baliyan2015review} and in some cases for various decades \cite{esteves2015long}. It confers month to month figure for top and valley in loads for different dissemination frameworks \cite{daneshi2008long}, on other hand, the short-term load forecasting covers time interims of another half hours to a week or some of the time few weeks [15]. It fulfils brief, termed targets like aggregated household demands, it is more accurate and smooth strategy utilized for load forecasting \cite{Jacob2020}.
\subsection{Recurrent Neural Networks}
\par RNN (Recurrent neural networks) is a machine learning approach that can remember the sequences of inputs, pattern recognition is the specialty of this technique, the processing time of RNN is much longer in comparison with ANN. Fig. X shows the working of a basic RNN cell.

\begin{figure} [h]
\hfil \includegraphics[width=7cm,height=4cm]{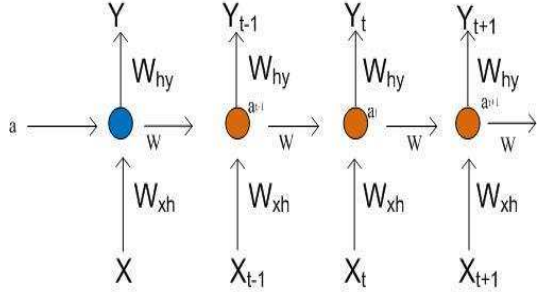}
\caption{Basic RNN}\label{fig:cm}
\end{figure}

Here, $W_{xh}$ is the weight for connection of the input layer to the hidden layer, $W$ is the weight for the connection of the hidden layer to the hidden layer, Why is the weight for the connection of the hidden layer \& $a$ is the activation layer.

\subsubsection{LSTM (Long Short Term Memory) Networks:}

\par LSTM is a recurrent neural network that consists of memory blocks in the recurrent hidden layer, this network contains multiplicative units called gates to control the flow of information in network. These have three gates comprising of input gates that controls the flow of input activations into the memory cell whereas output gates control the output flow of cell activations into the rest of the network on the other hand there is one gate known as forget gate that scales the internal state of the cell.

\par LSTM uses no activation function within its recurrent components and it is capable of remembering values for either long or short time periods. Stack LSTM model is the one of the models in which layers are stacked, the stacking of three LSTM forms a deep RNN. \figurename{ \ref{fig:LSTM}} shows the internal structure of the LSTM cell.

\begin{equation}
f_t = \sigma(W_f\cdot[h_{t-1}, x_t] + b_t)
\end{equation}
\begin{equation}
i_t = \sigma(W_i\cdot[h_{t-1}, x_t] + b_i])
\end{equation}
\begin{equation}
\tilde{C_t} = tanh(W_c\cdot[h_{t-1}, x_t] + b_t)
\end{equation}
\begin{equation}
C_t = f_t\cdot C_{t-1} + i_t \cdot \tilde{C_t}
\end{equation}
\begin{equation}
o_t = \sigma(W_o\cdot[h_{t-1}, x_t] + b_o)
\end{equation}
\begin{equation}
h_t = o_t\cdot tanh(C_t)
\end{equation}

\par Where, $W$ is the weight matrices, $C_t$ is the cell state and $b$ is the input bias vector whereas $i$, $f$, $o$ is the input, forget and the output gate layer respectively. Cell out activation function in this paper is $tanh$.

\begin{figure} [h!]
\begin{minipage}{0.48\linewidth}
\includegraphics[width=\linewidth]{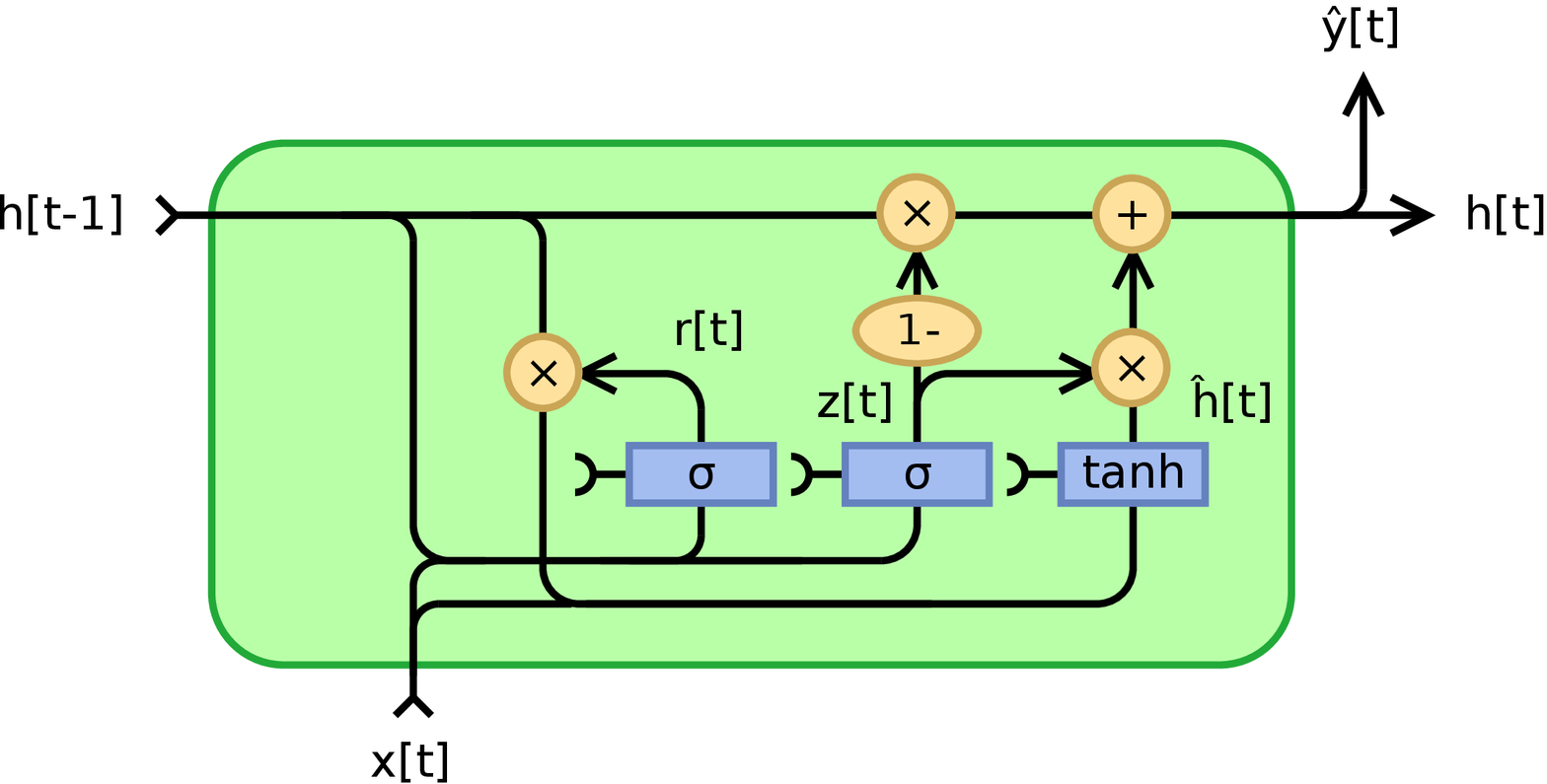}
\caption{LSTM cell}\label{fig:LSTM}
\end{minipage}
\begin{minipage}{0.48\linewidth}
\includegraphics[width=\linewidth]{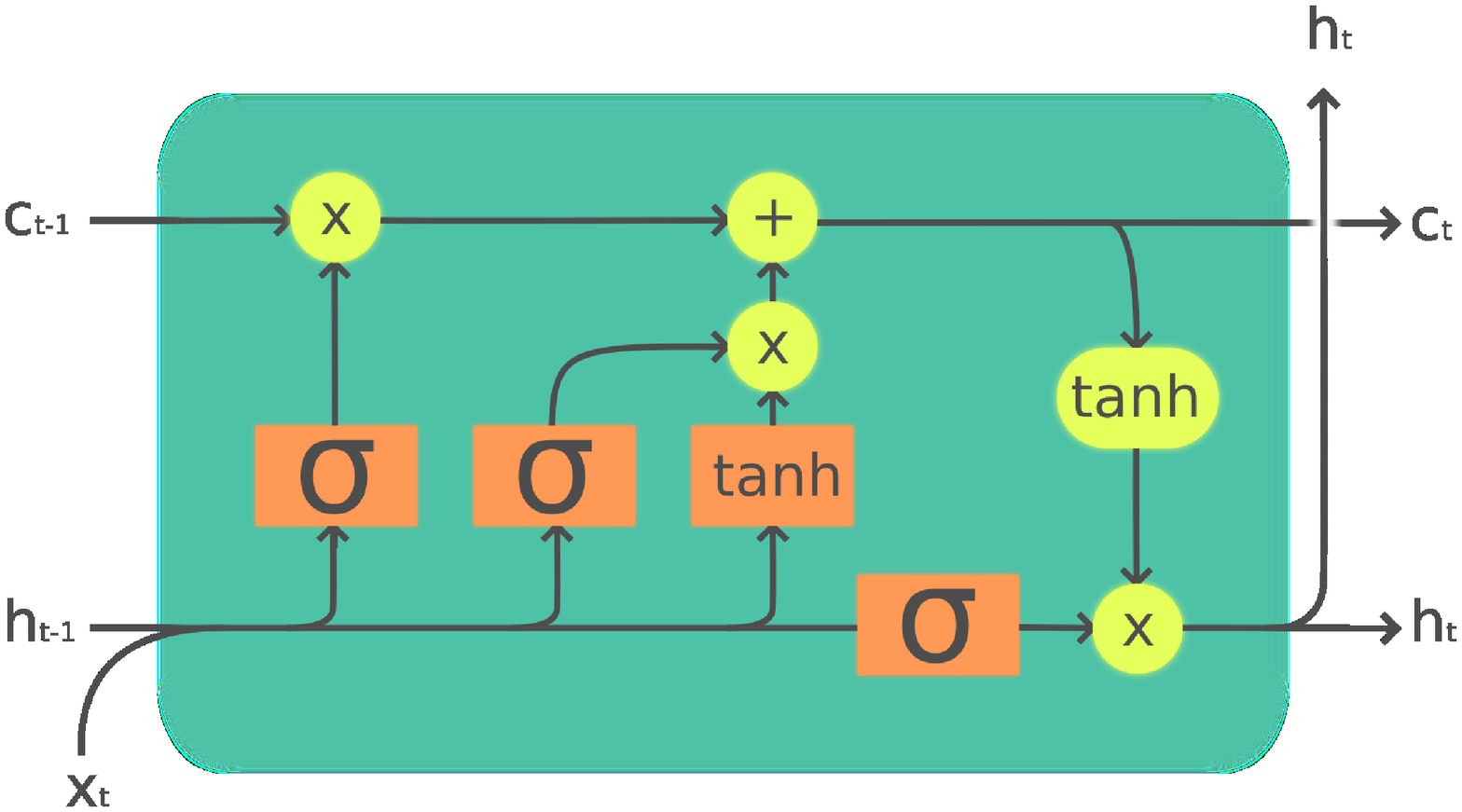}
\caption{GRU cell}\label{fig:GRU}
\end{minipage}
\end{figure}

\subsubsection{GRU (Gated Recurrent Unit) Networks:}

\par Unlike other variants of LSTM, GRU is relatively easier to train and it is little faster than the traditional LSTM. GRU combines the gating functions of input gate j and forget gate f into a single update gate basically the cell state positions are marked by entry points for new data. GRU is used for the exotic concepts like neural GPUs[2].

\begin{equation}
h_t = (1 - z_t)\cdot h_{t-1} + z_t\cdot \tilde{h_t}
\end{equation}
\begin{equation}
\tilde{h_t} = g(w_hx_t + U_h(r_t\cdot h_{t-1})) + b_h
\end{equation}

\subsubsection{Convolution LSTM}
\par An approach of a ConvLSTM could be a Convolution-LSTM model, in which the picture passes through the convolutions layers and its result is a set smoothed to a 1D cluster with the gotten features. It is the matrix multiplication calculation of the input with the LSTM cell supplanted by the convolution operation. The output of the convolutional layer as the input to this layer. The LSTM model contains multiple LSTM cells, where each LSTM cell consists of an input gate ($i_t$), forget gate ($f_t$) and output gate ($o_t$). The calculation is as follows:

\begin{equation}
f_t	 = \sigma(W_f\cdot[h_{t-1}, y_t] + b_f)
\end{equation}
\begin{equation}
i_t = \sigma(W_i\cdot[h_{t-1}, y_t] + b_i])
\end{equation}
\begin{equation}
o_t = \sigma(W_o\cdot[h_{t-1}, y_t] + b_o)
\end{equation}
\begin{equation}
g_t = \sigma(W_g\cdot[h_{t-1}, y_t] + b_g)
\end{equation}
\begin{equation}
C_t = f_t\odot C_{t-1} + i_t\odot g_t
\end{equation}
\begin{equation}
h_t = o_t\odot tanh(C_t)
\end{equation}

where, $\odot$ represents element-wise multiplication, $\sigma$ denotes the sigmoid function containing the gating values in $[0, 1]$. $y_t$ $\in$ $R_q$ is the current input from the lower layer at time step t (where $q$ is the dimensionality of word vector $y_t$). $H_{t-1}$ is the output vector of the previous step, and $W_i$ , $b_i$ , $W_f$ , $b_f$ , $W_o$, $b_o$, $W_g$ and $b_g$ are the parameters that must be trained[10].
\begin{figure} [h!]
\hfil \includegraphics[width=14cm ,height=8.5cm]{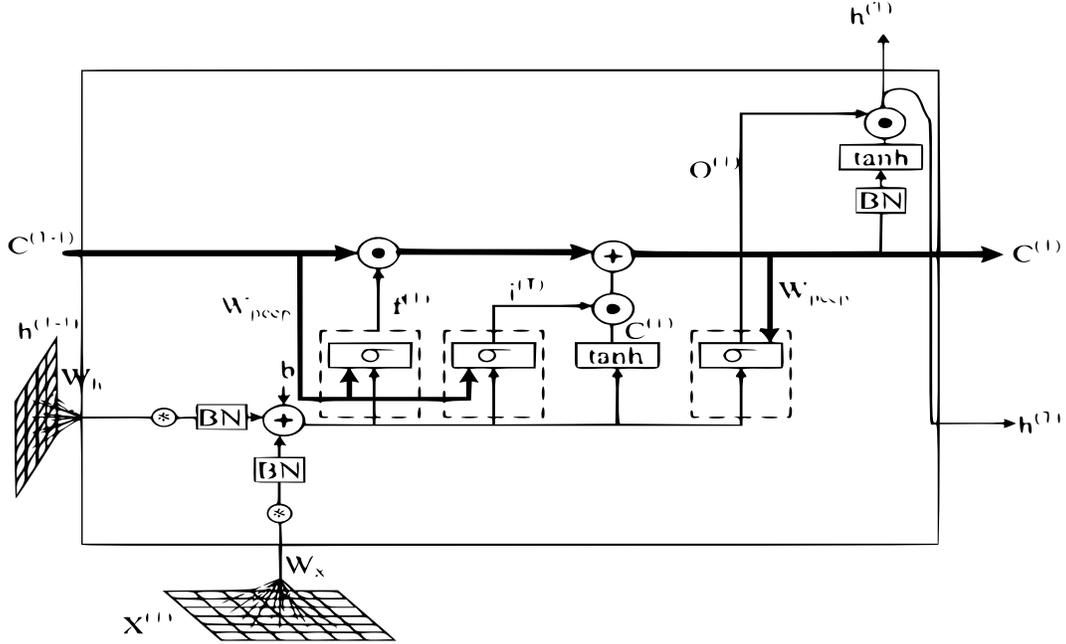}
\caption{Convolution LSTM cell\cite{kadupitigesurvey}}\label{fig:ConvLSTM}
\end{figure}
\subsubsection{Bidirectional LSTM Networks:}
\par Bidirectional LSTMs are an expansion of conventional LSTMs that can progress model execution on arrangement classification problems. In problems where all timesteps of the input arrangement are accessible. Bidirectional LSTMs train two rather than one LSTMs on the input arrangement. The primary on the input sequence as-is and the second on a turned around duplicate of the input sequence. This could give extra setting to the organize and result in speedier and indeed more full learning on the issue [9].
\begin{figure} [h!]
\includegraphics[width=\linewidth ,height=7cm]{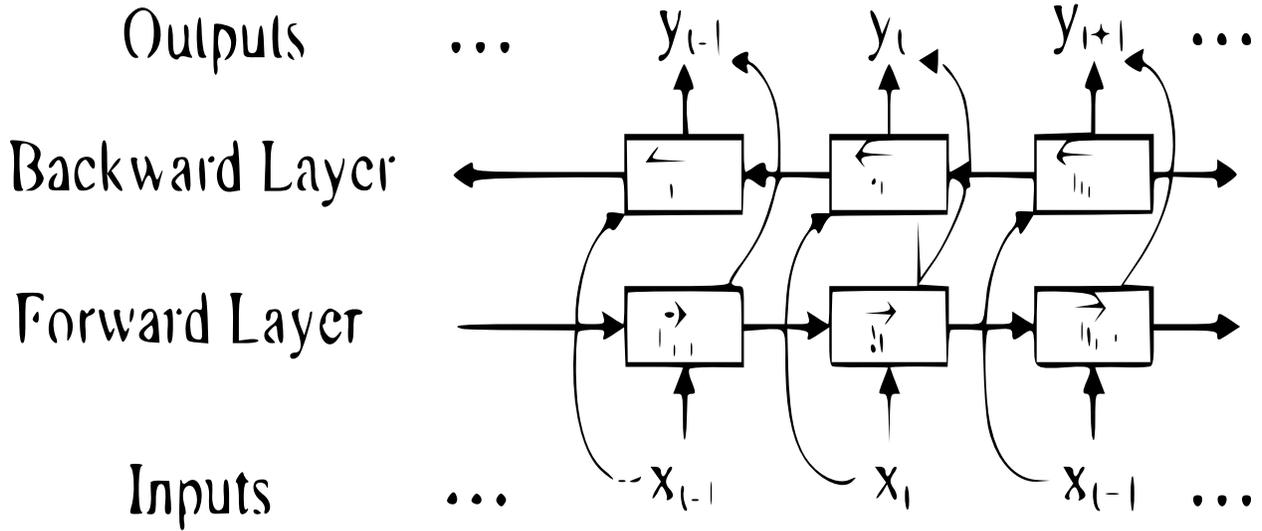}
\caption{Bidirectional LSTM cell}\label{fig:LSTM}
\end{figure}
\subsection{Dataset used} 
\par The data used in this paper was collected and prepared by Enerdata organization \cite{dsener}. The data originally contains yearly electricity domestic consumption (in $TWh$) for $61$ entities (including countries, continents and unions) from the year $1990$ to $2019$. This paper focuses mainly on G-20 members, which includes : Argentina, Australia, Brazil, Canada, China, France, Germany, India, Indonesia, Italy, Japan, South Korea, Mexico, Russia, Saudi Arabia, South Africa, Turkey, United Kingdom, United States, European Union. Among all G-20 members, $19$ are the countries and one is the European Union.

\section{Results and Discussions}

\par
\begin{figure}[!h]
\hfil \includegraphics[width=13cm,height=7cm]{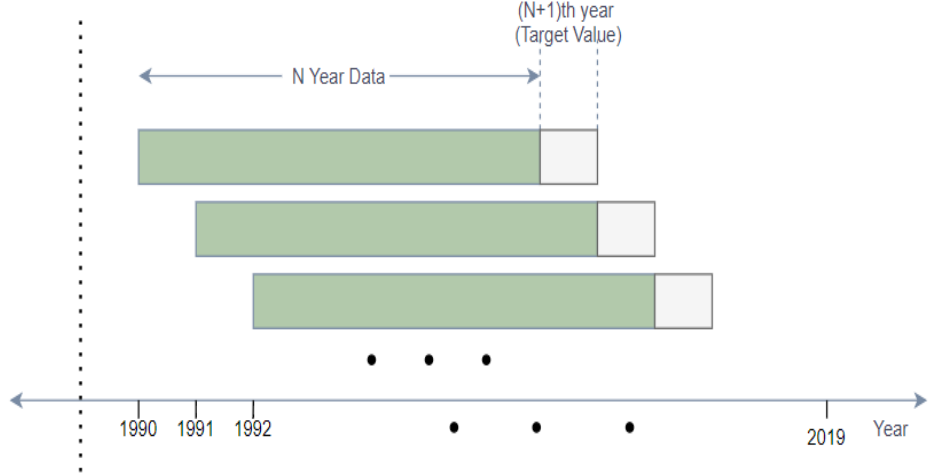}
\caption{Demonstration of Sliding Window approach\label{win}}
\end{figure}

 Recurrent neural networks can process the data sequence of any arbitrary size. In light of the fact that we have less data, sliding window approach can be used to train the recurrent model. Sliding window approach provides $N$ timesteps and expect the model to predict $(N +1)^{th}$ timestep. This approach can give us large amount of training data compared to raw training data at the cost of data repetition. This way model can learn trend from previous values in the chosen window size, for instance window size 4 will give previous 4 years data and will expect the model to predict value of the $5^{th}$ year. Fig. \ref{win} shows the demonstration of sliding window on the time-series data. As seen in the figure, $N$ values are given as an input to the model while, $(N + 1)^{th}$ value is the target value. Window is shifted one by one after creating each training instance. We created the new datasets with various window sizes in the range $3-7$ while, keeping the data of the years $2016-2019$ reserved for the testing purpose.

\par In addition to different windows sizes, we also experimented with LSTM, GRU, Bidirectional-LSTM and Conv-LSTM. Models based on LSTM, GRU and Bidirectional-LSTM has two recurrent layers stacked each with $36$ units followed by a dense layer. First two layers have $ReLU$ as an activation function while, last has linear activation function which simply gives the activation values without applying any additional computation. Conv-LSTM based model has similar architecture except it has $64$ filters in first two layers followed by the flatten layer which simply flattens the previous layer's output so, it can be fed to the last dense layer. All of the models were trained for the maximum of $200$ epochs with $Adam$\cite{kingma2014adam} as an optimizer (having the slow learning rate of $0.001$) and $huber$ as the loss function. Fig. \ref{we} shows the MAE (Mean Absolute Error) achieved by each of the models for the different window sizes for the test data. 

\begin{figure}
\begin{center}

	\begin{tabular}{p{2cm}  p{2cm}  p{3cm}  p{2cm}  p{2.5cm} }
	\toprule
	 \hfil Window &\multicolumn{4}{c}{Mean Absolute error(MAE) in TWh}	\\
	\cline{2-5}
    \hfil Size & LSTM & Bidirectional LSTM & GRU & ConvLSTM\\
    \midrule
    \hfil 3 & $21.646873$ & \hfil $20.861853$ & $19.607866$ & $19.545732$ \\
    \hfil 4 & $19.642729$ &\hfil $20.14976$ & $19.1428$ & $19.634954$ \\
    \hfil 5 & $20.922813$ & \hfil $19.376783$ & $19.574736$ & $20.03209$ \\
    \hfil 6 & $\textbf{16.2193}$ & \hfil $17.725124$ & $18.281723$ & $17.696333$ \\
    \hfil 7 & $19.55587$ & \hfil $20.1838$ & $19.3956$ &  $17.649738$\\
    \bottomrule
	\end{tabular} 
\caption{MAE on test data of each trained model\label{we}}

\end{center}

\end{figure}
\begin{figure}[!h]
\begin{subfigure}[h]{\linewidth}
\includegraphics[width=5.2cm,height=4.1cm]{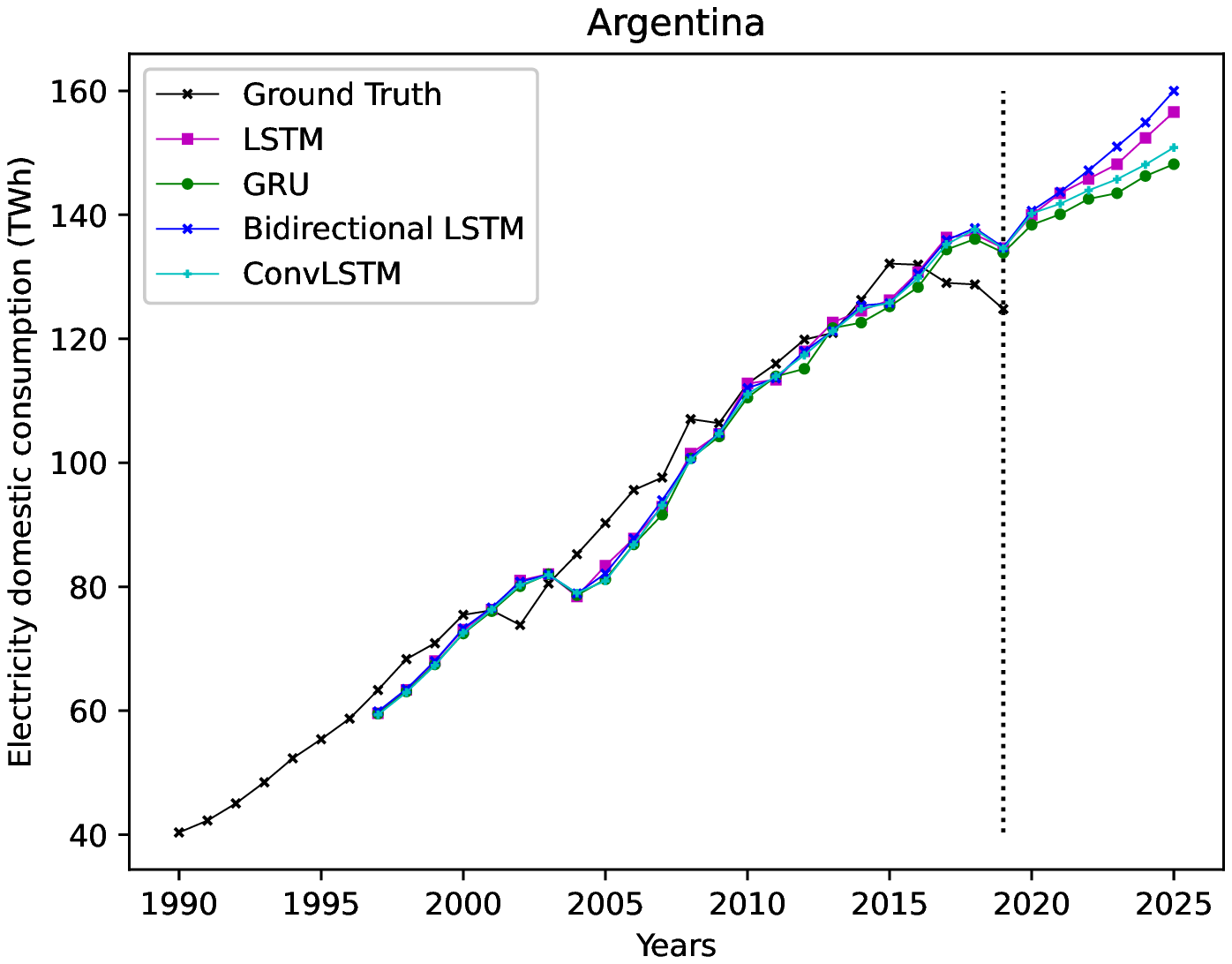}
\includegraphics[width=5.2cm,height=4.1cm]{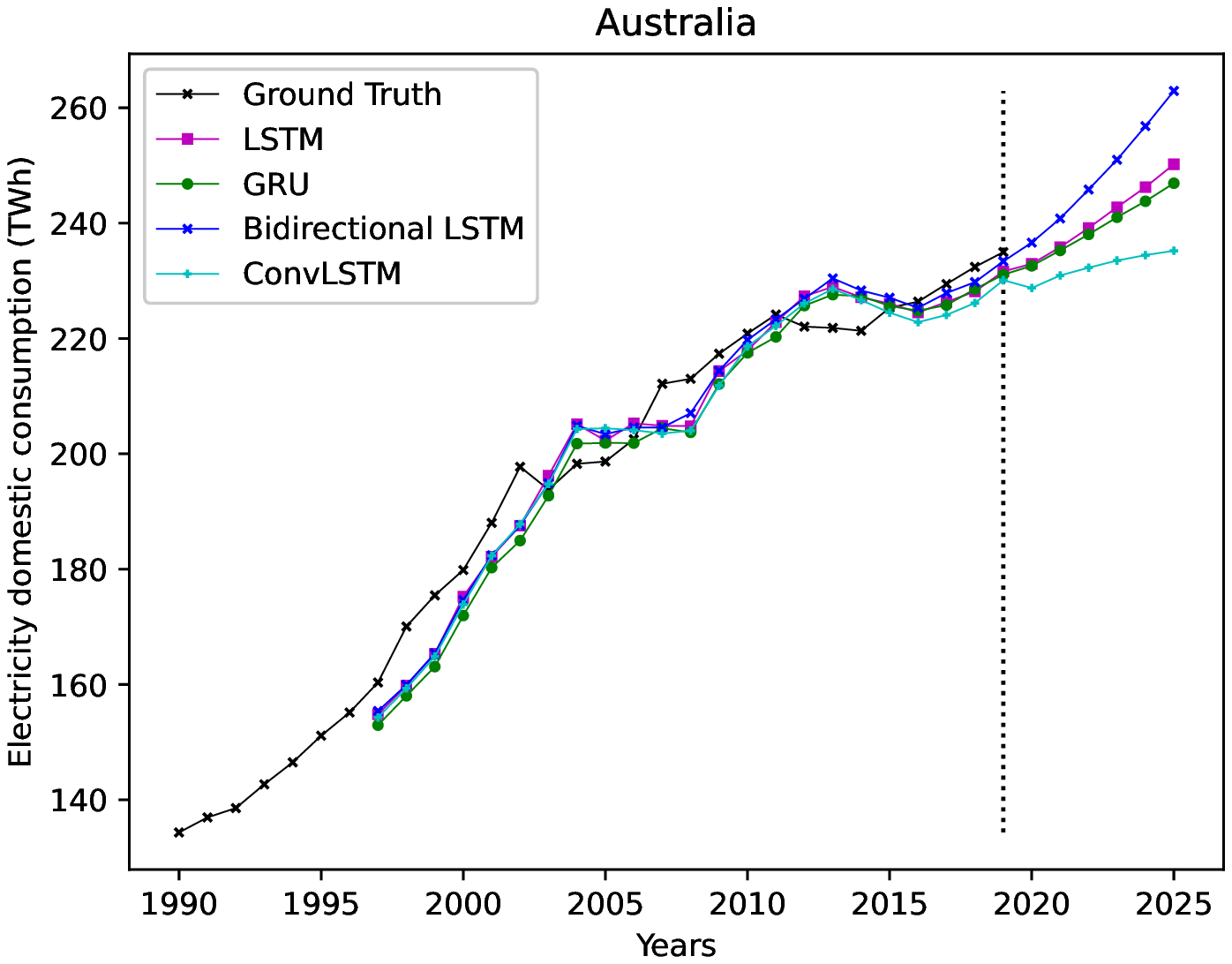}
\includegraphics[width=5.2cm,height=4.1cm]{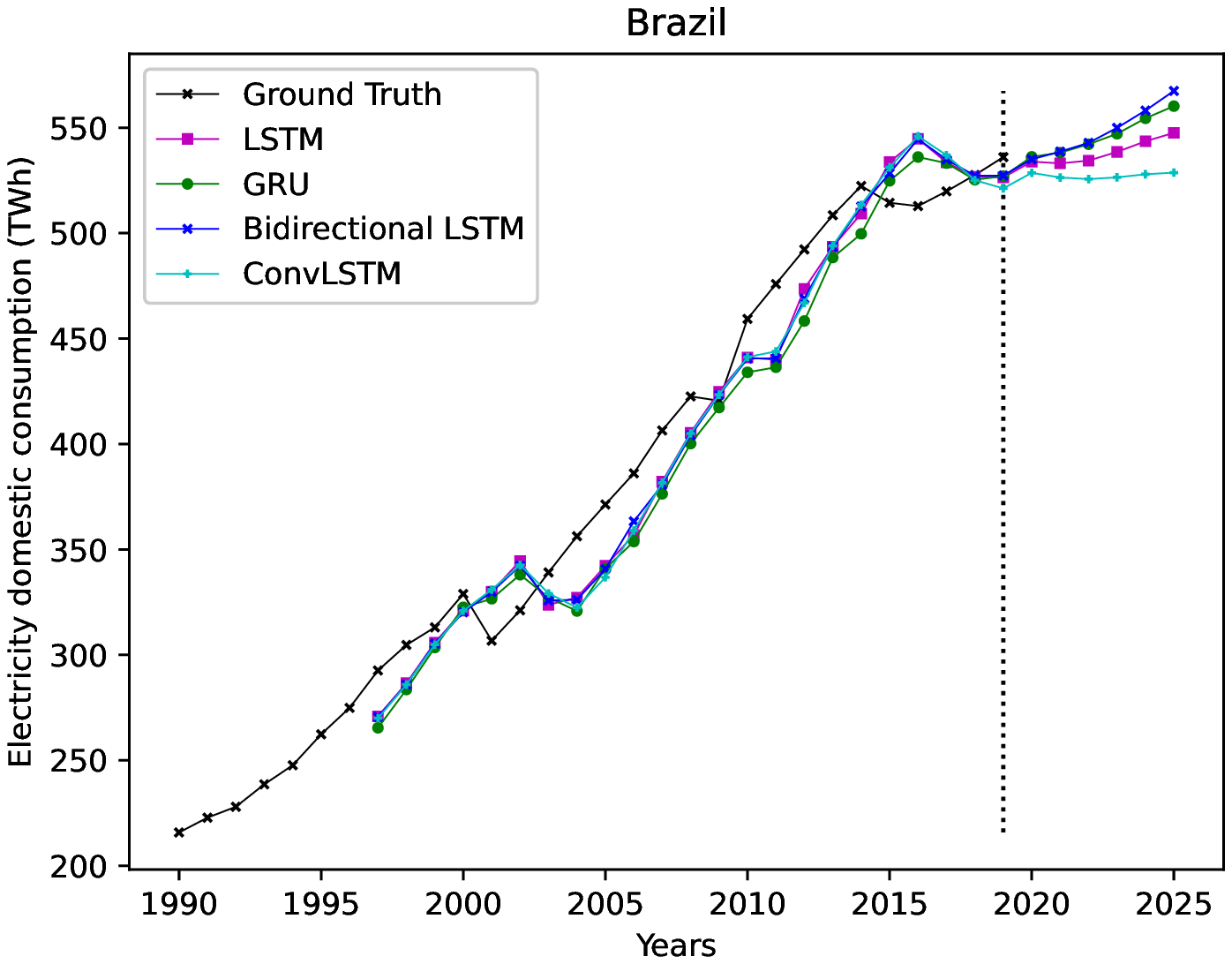}
\end{subfigure}
\end{figure}
\begin{figure}[!h]
\begin{subfigure}[h]{\linewidth}
\includegraphics[width=5.2cm,height=4.1cm]{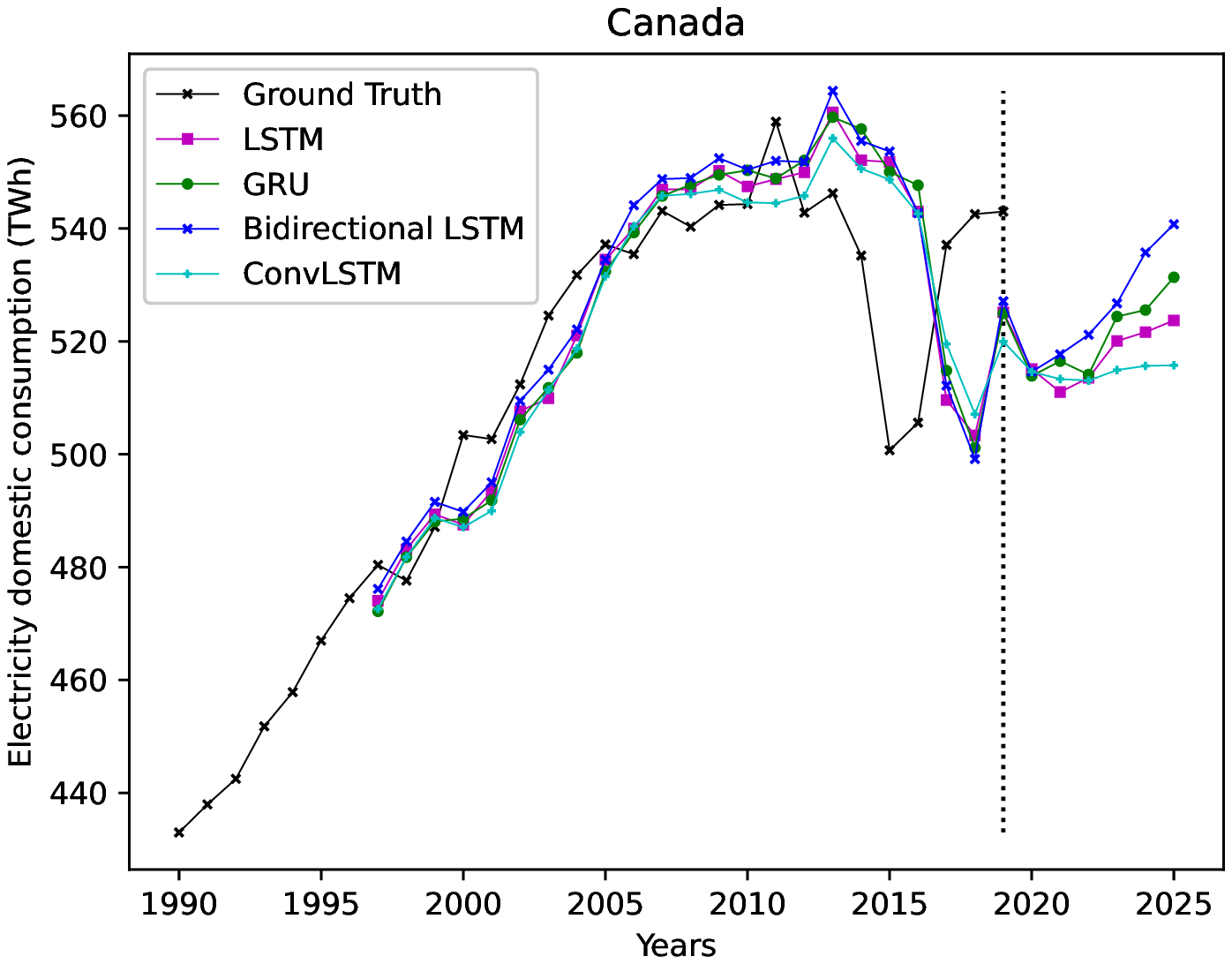}
\includegraphics[width=5.2cm,height=4.1cm]{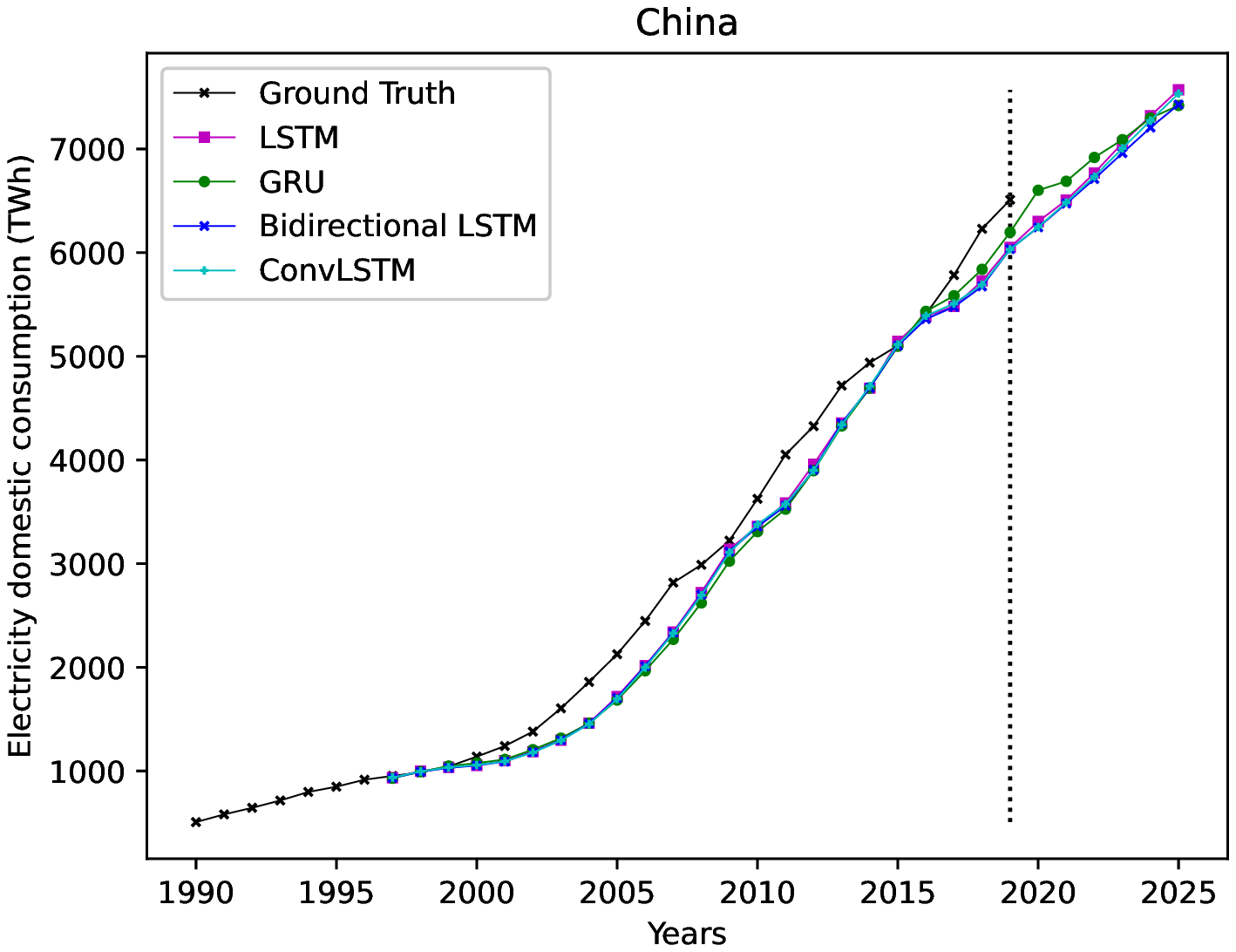}
\includegraphics[width=5.2cm,height=4.1cm]{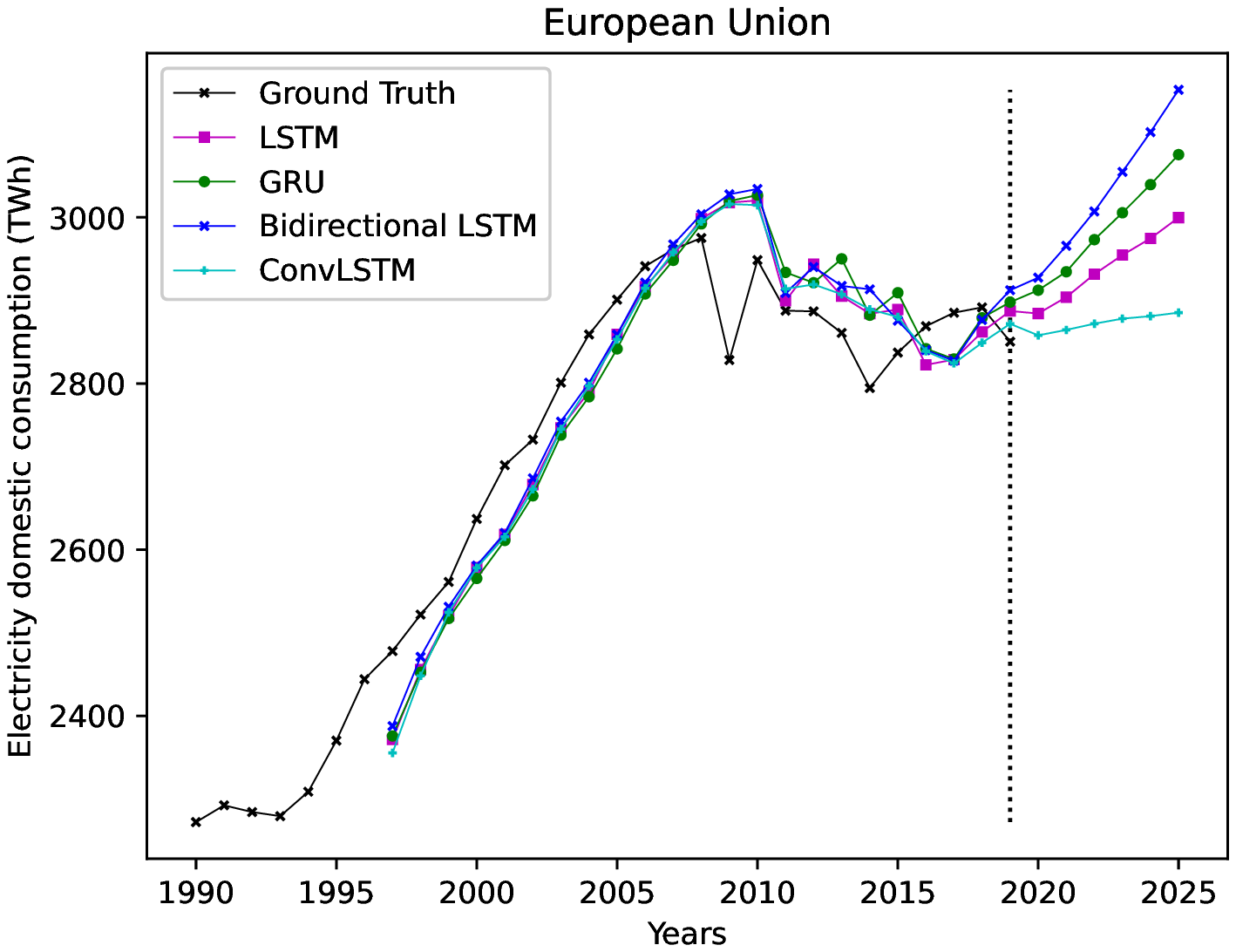}
\end{subfigure}
\end{figure}
\begin{figure}[!h]
\begin{subfigure}[h]{\linewidth}
\includegraphics[width=5.2cm,height=4.1cm]{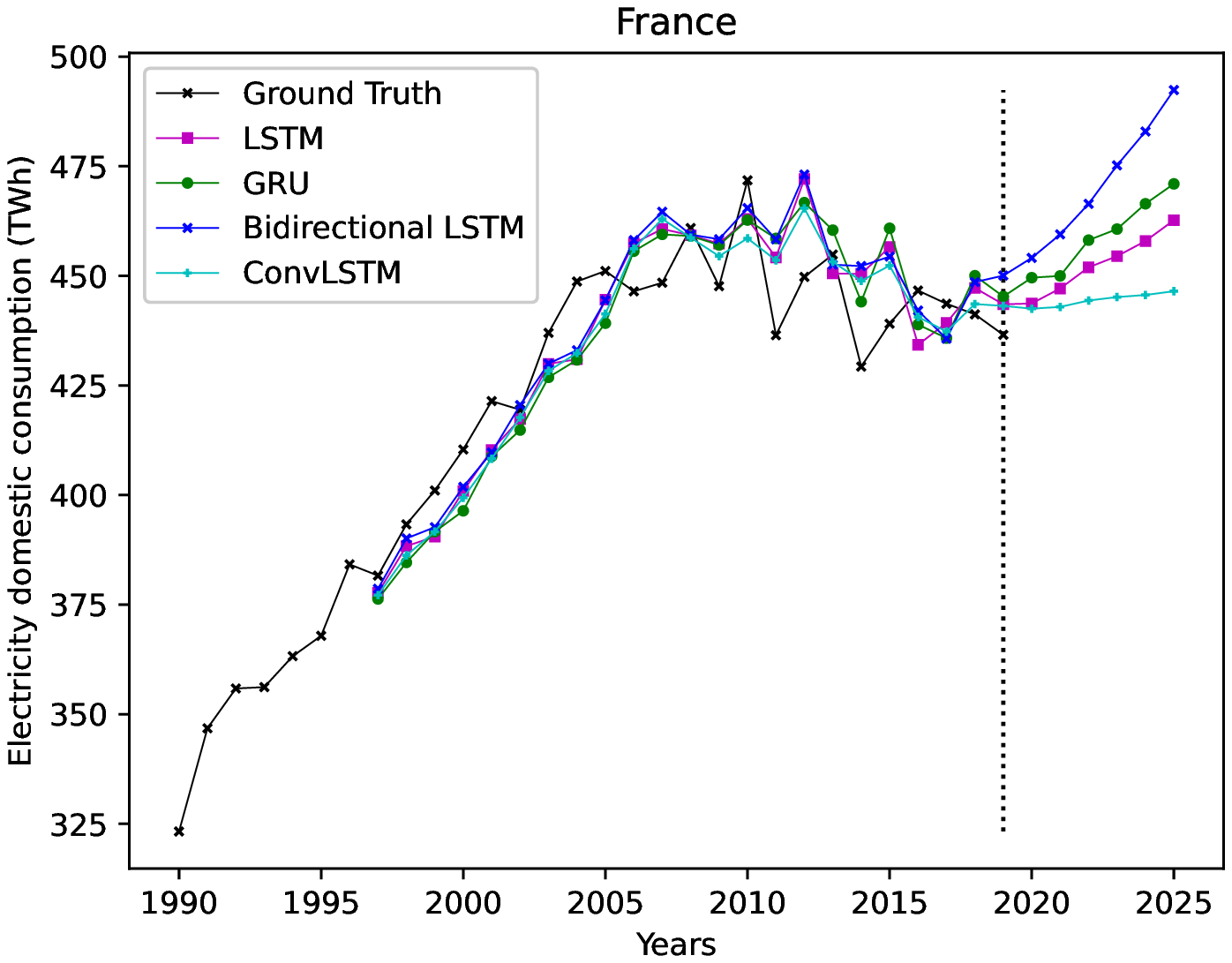}
\includegraphics[width=5.2cm,height=4.1cm]{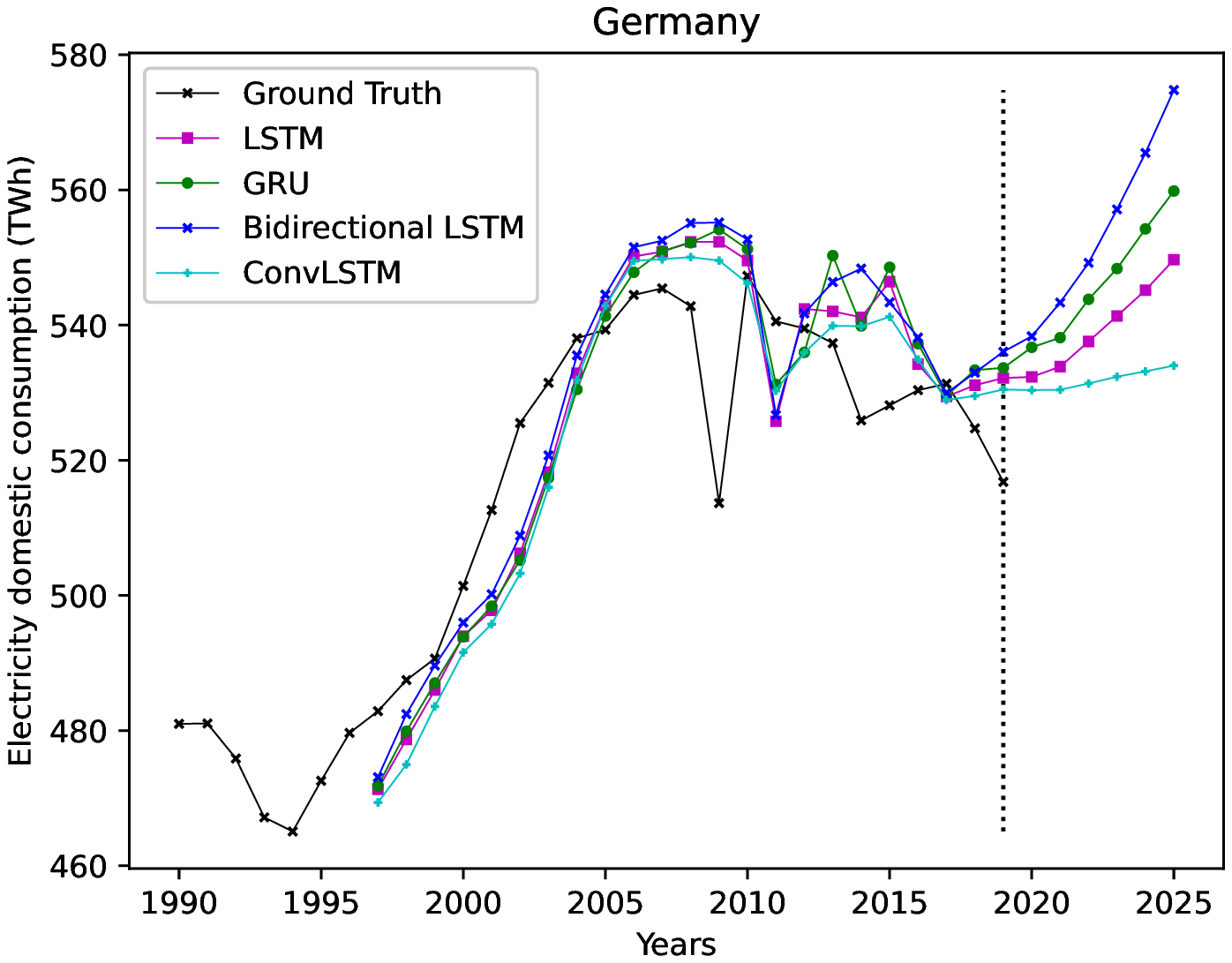}
\includegraphics[width=5.2cm,height=4.1cm]{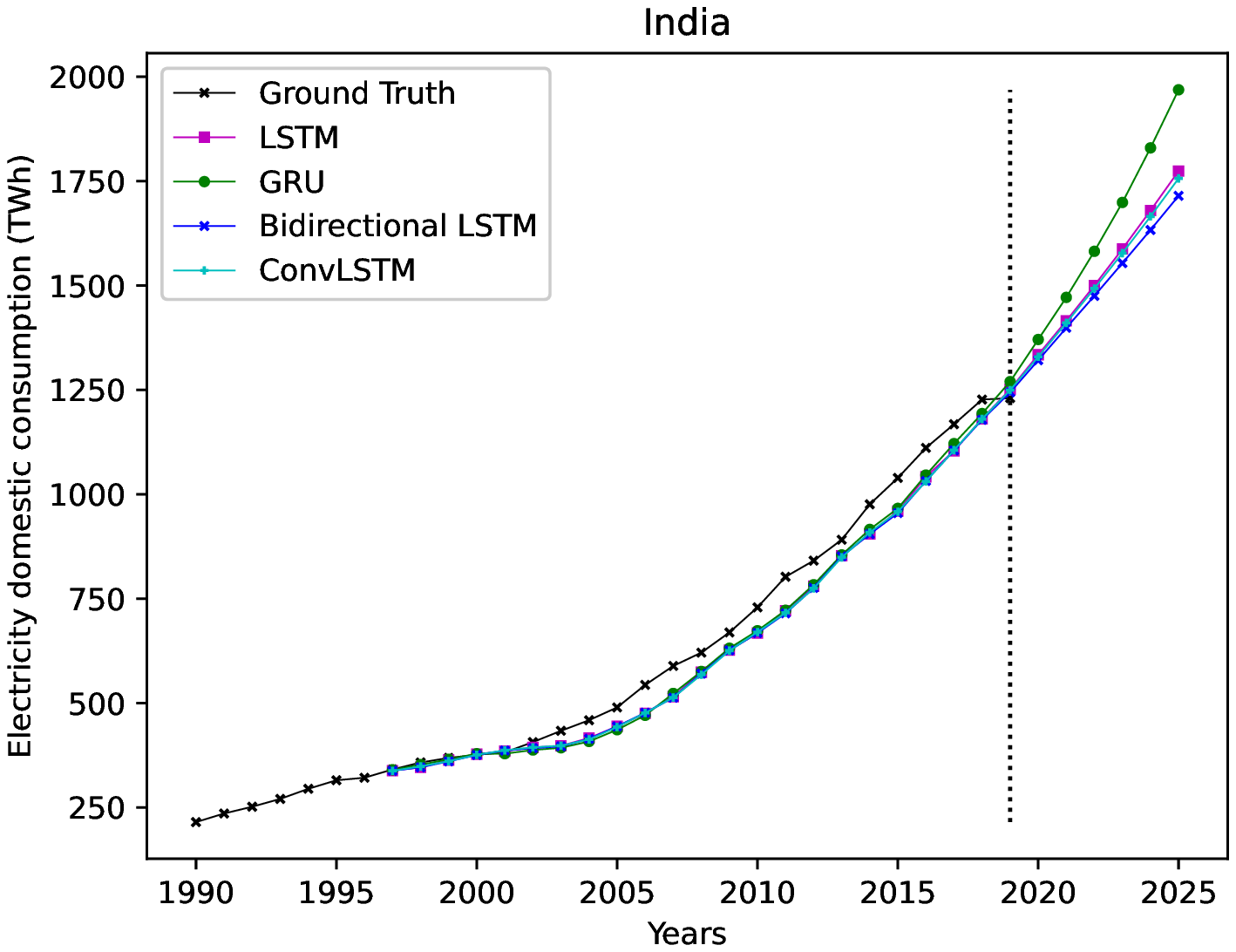}
\end{subfigure}
\end{figure}
\begin{figure}[!h]
\begin{subfigure}[h]{\linewidth}
\includegraphics[width=5.2cm,height=4.1cm]{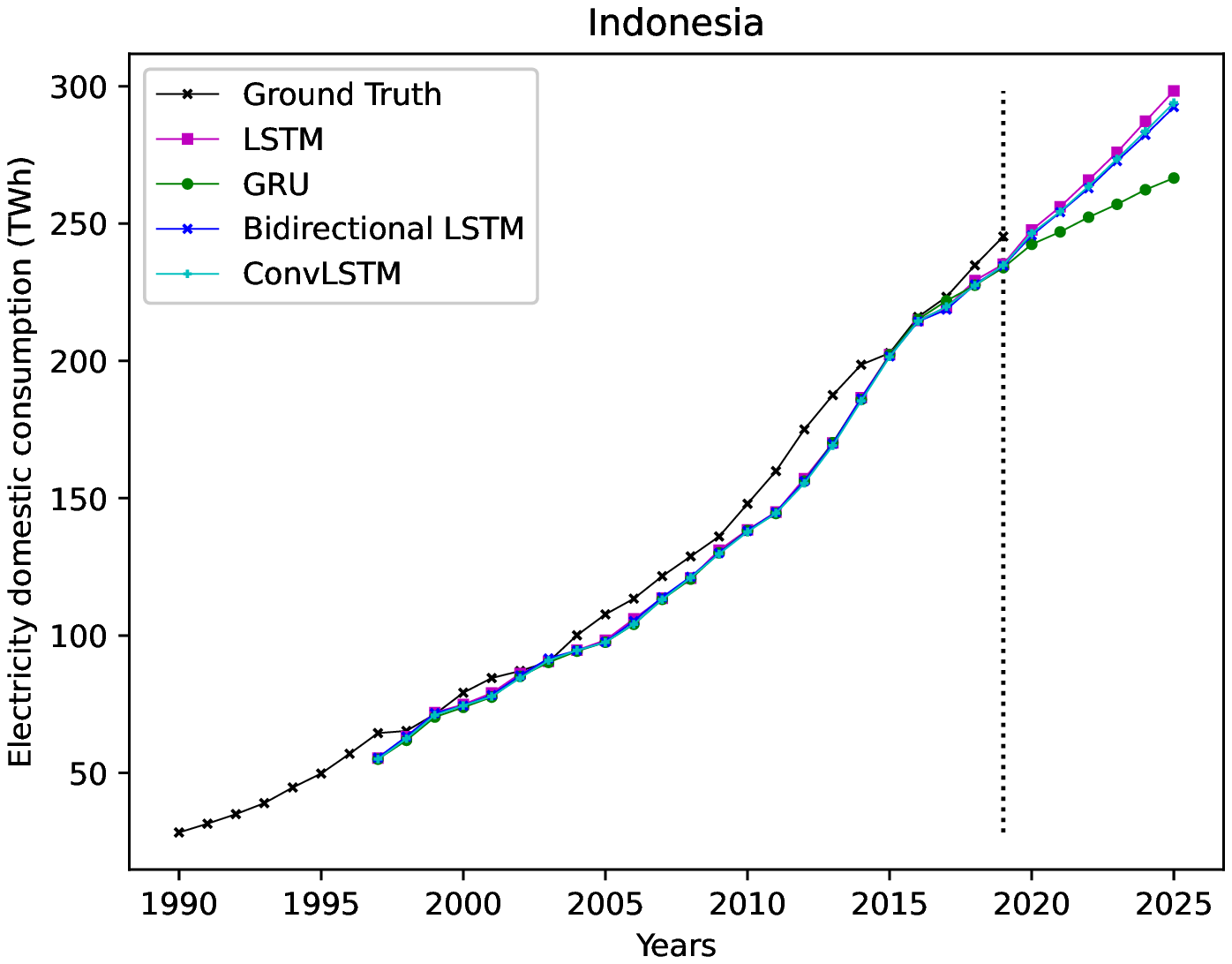}
\includegraphics[width=5.2cm,height=4.1cm]{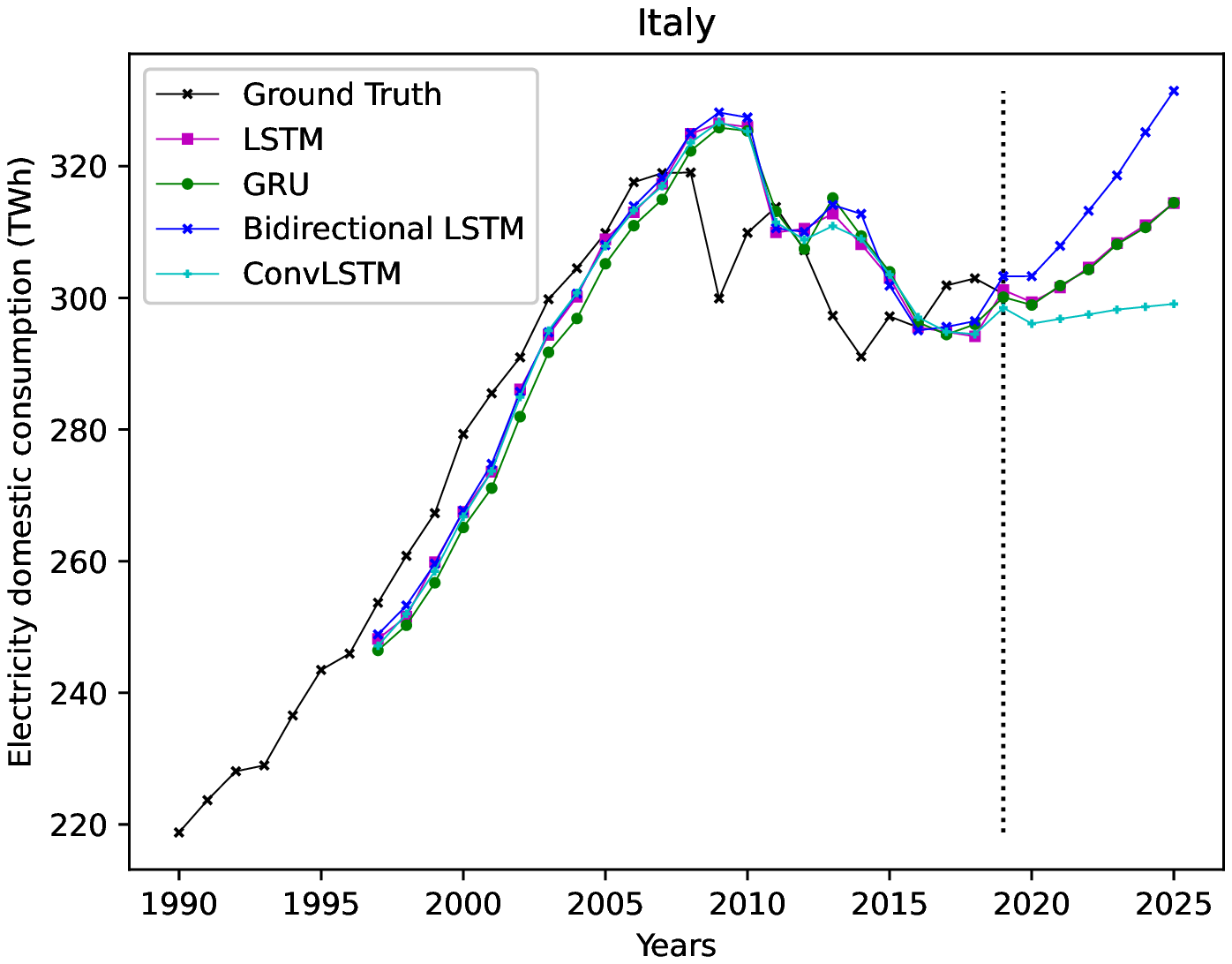}
\includegraphics[width=5.2cm,height=4.1cm]{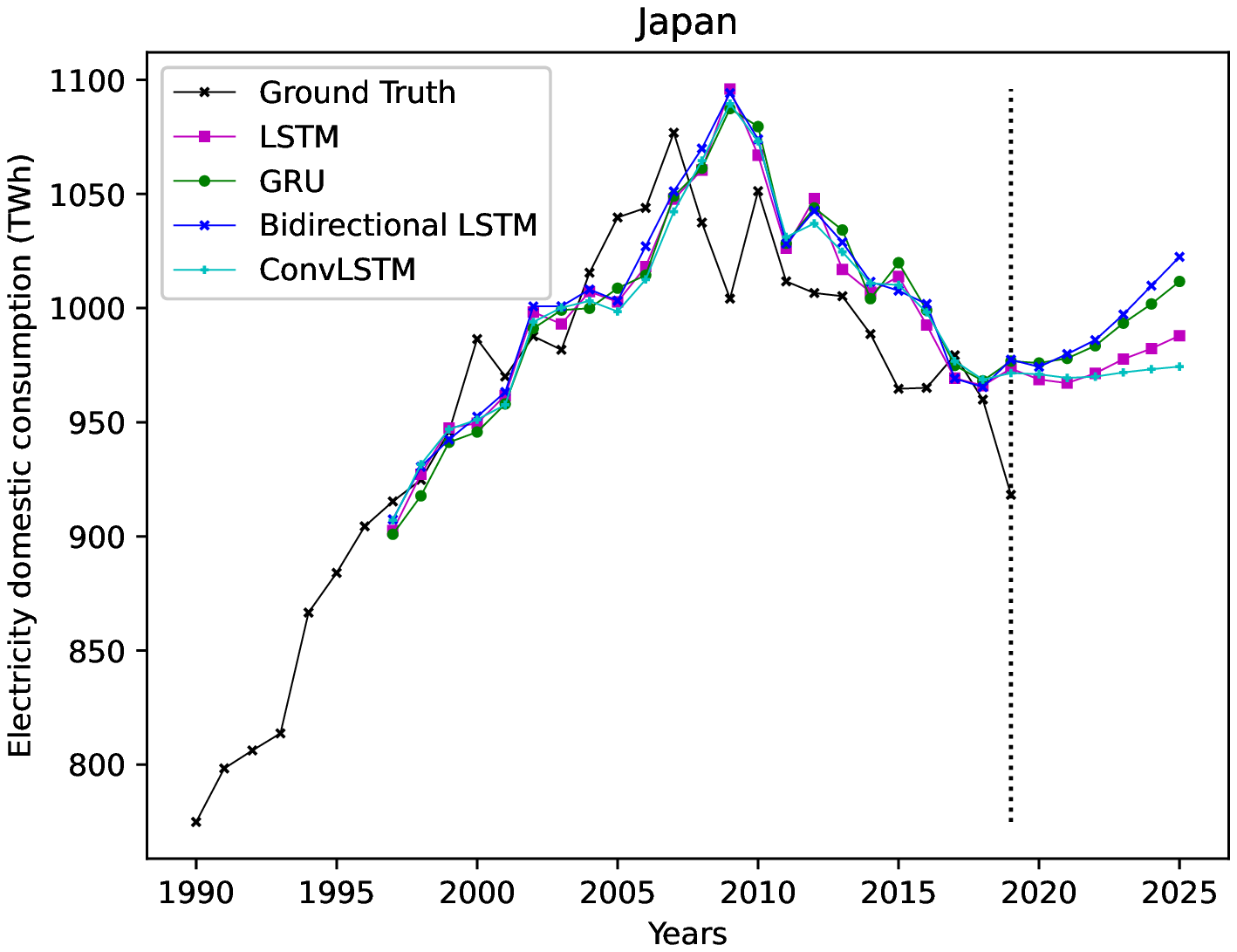}
\end{subfigure}
\end{figure}
\begin{figure}[!h]
\begin{subfigure}[h]{\linewidth}
\includegraphics[width=5.2cm,height=4cm]{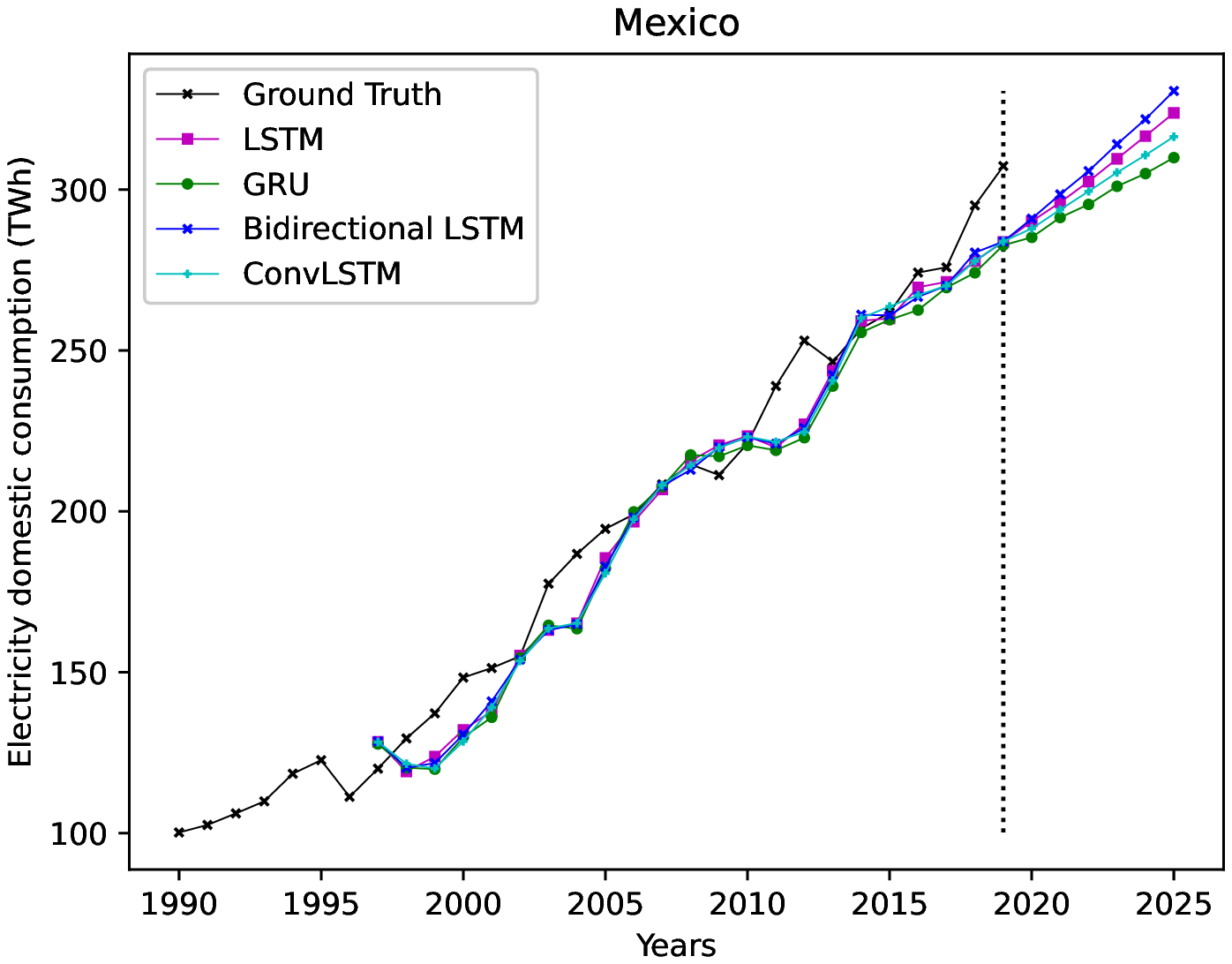}
\includegraphics[width=5.2cm,height=4cm]{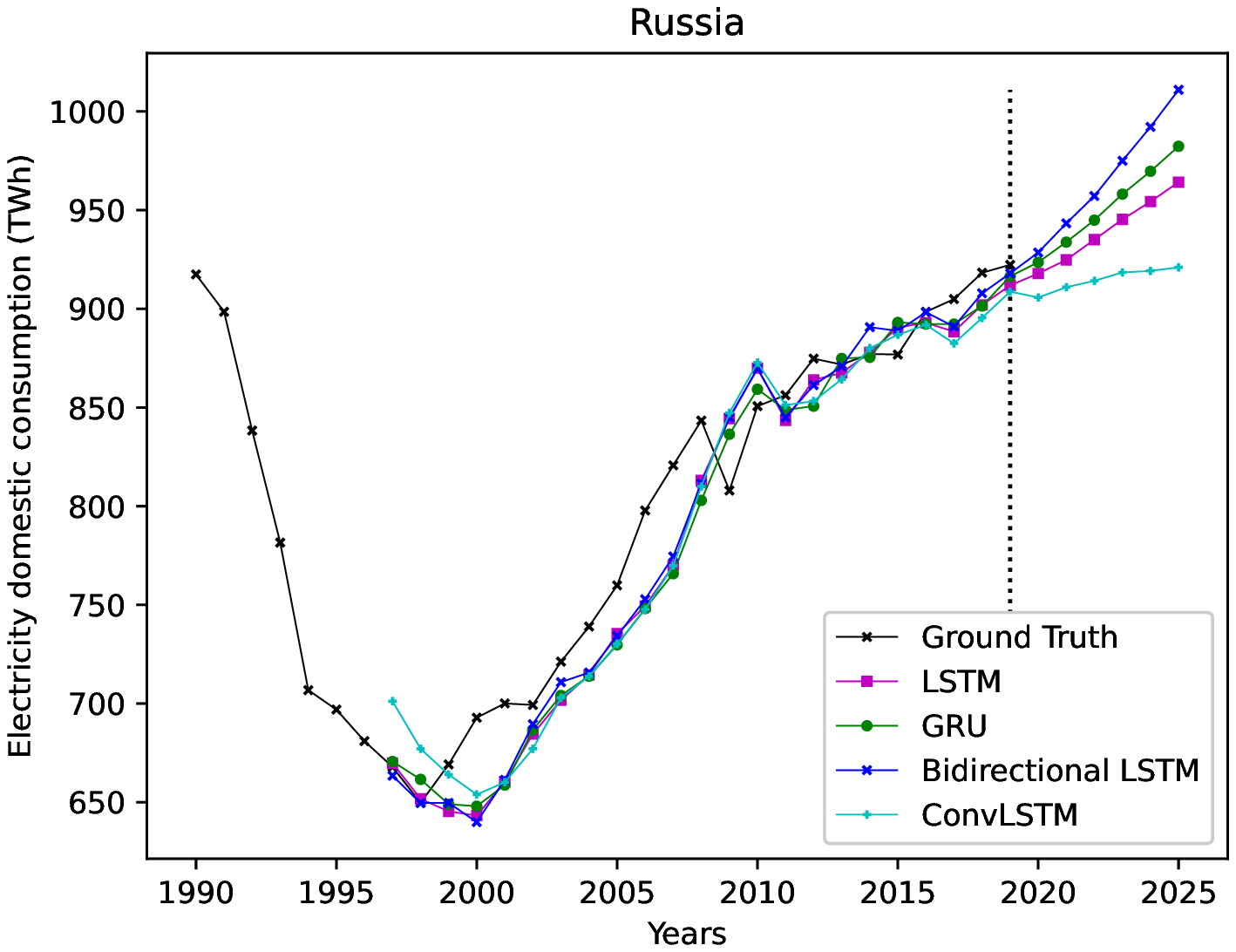}
\includegraphics[width=5.2cm,height=4cm]{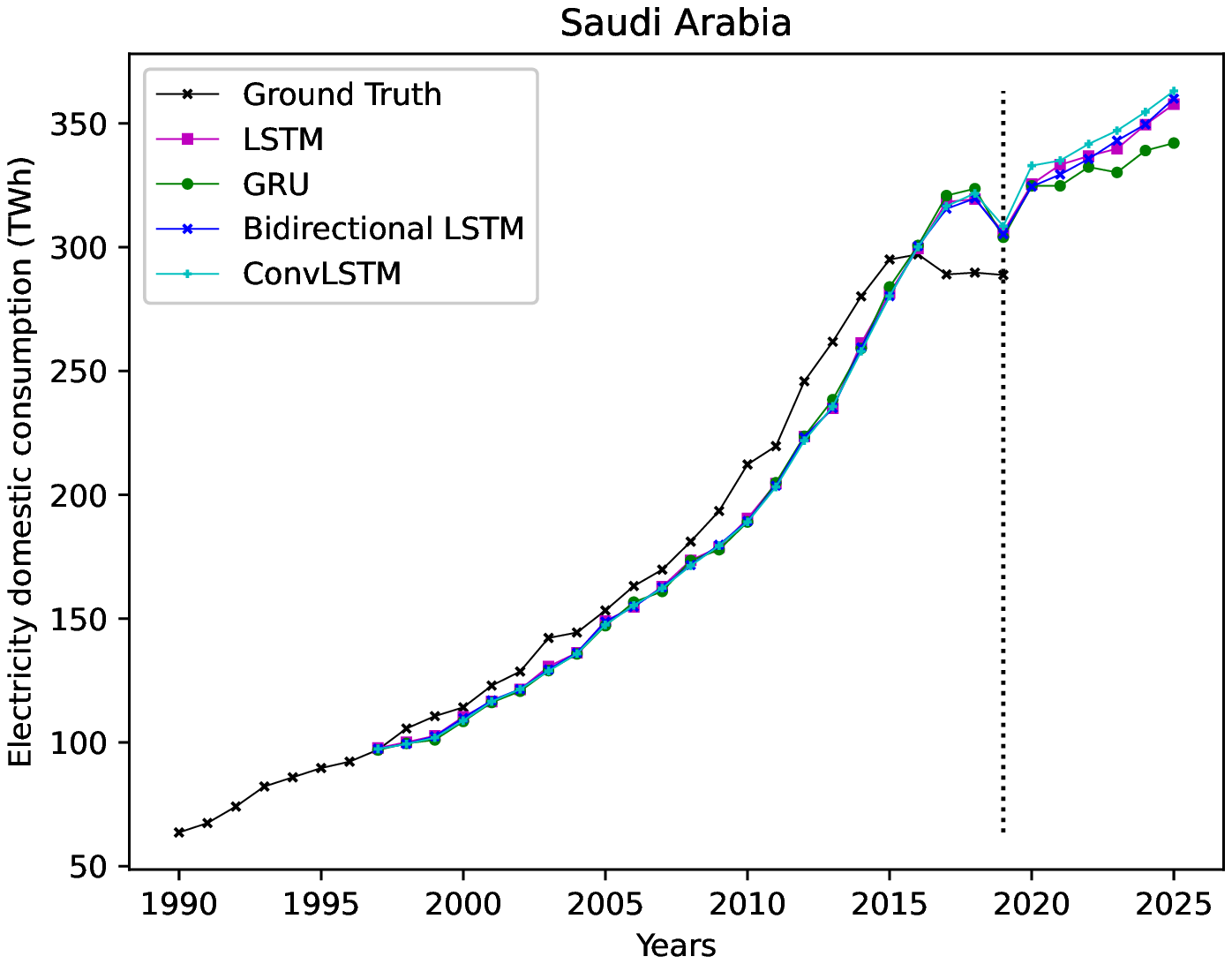}
\end{subfigure}
\end{figure}
\begin{figure}
\begin{subfigure}[h]{\linewidth}
\includegraphics[width=5.2cm,height=4.2cm]{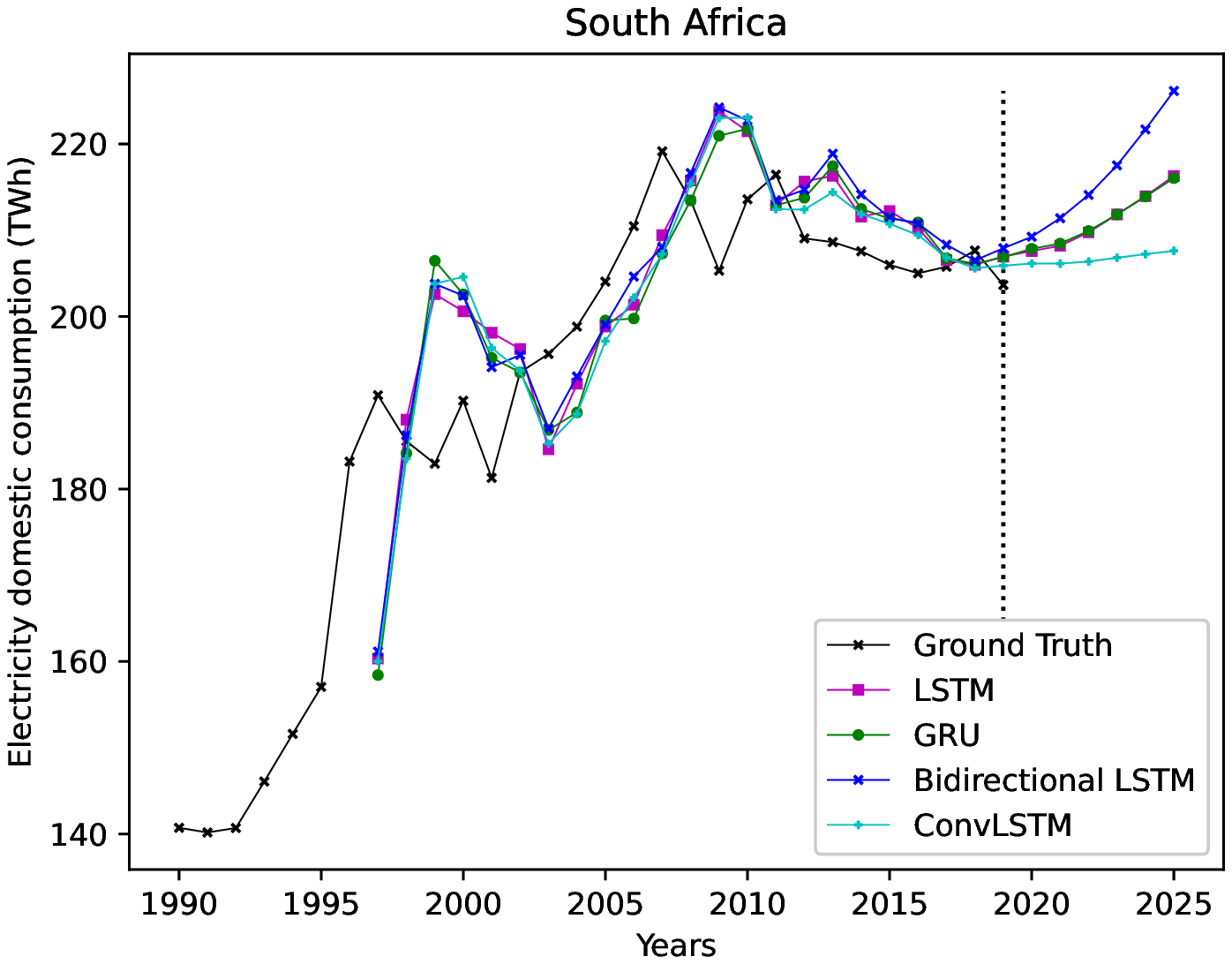}
\includegraphics[width=5.2cm,height=4.2cm]{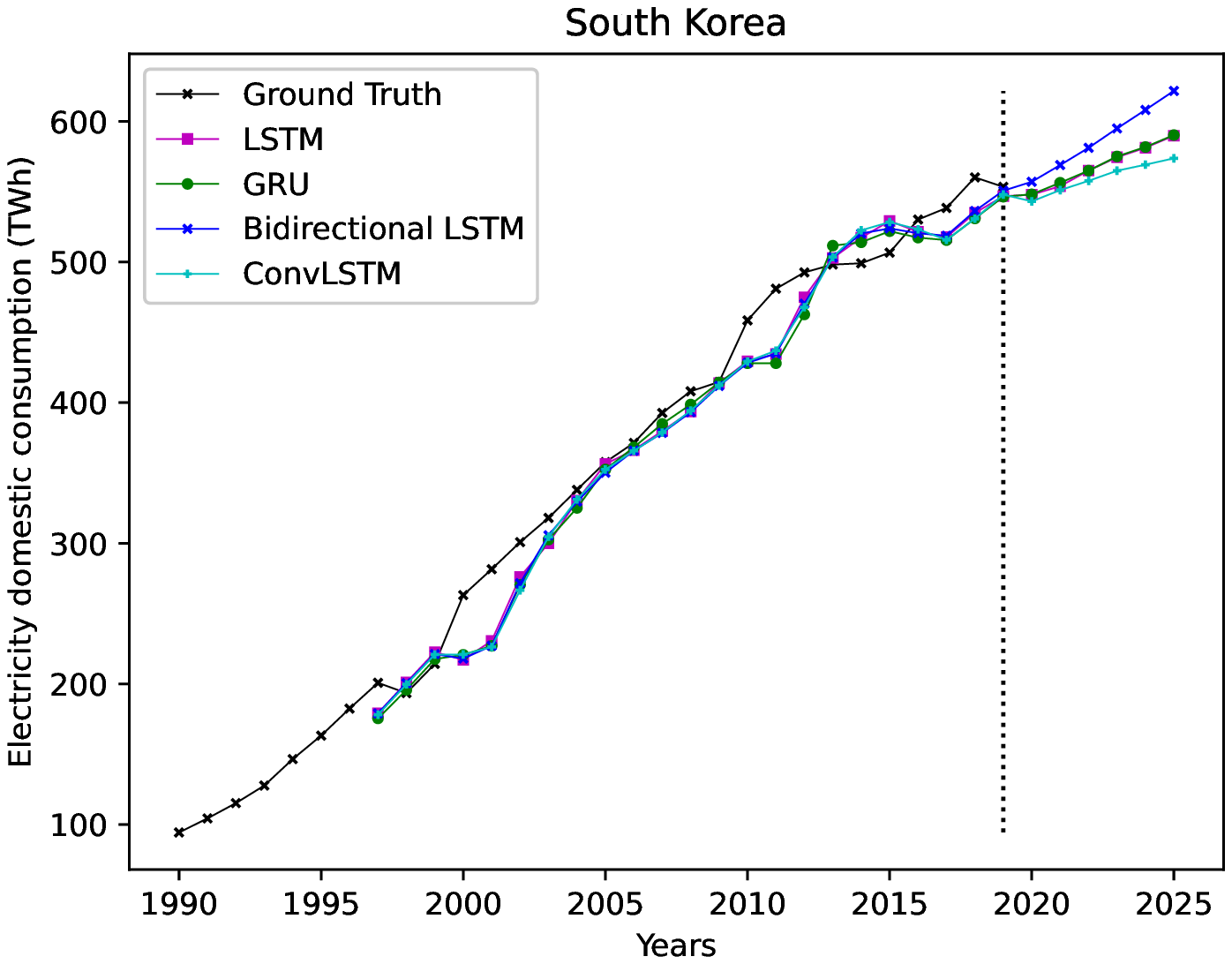}
\includegraphics[width=5.2cm,height=4.2cm]{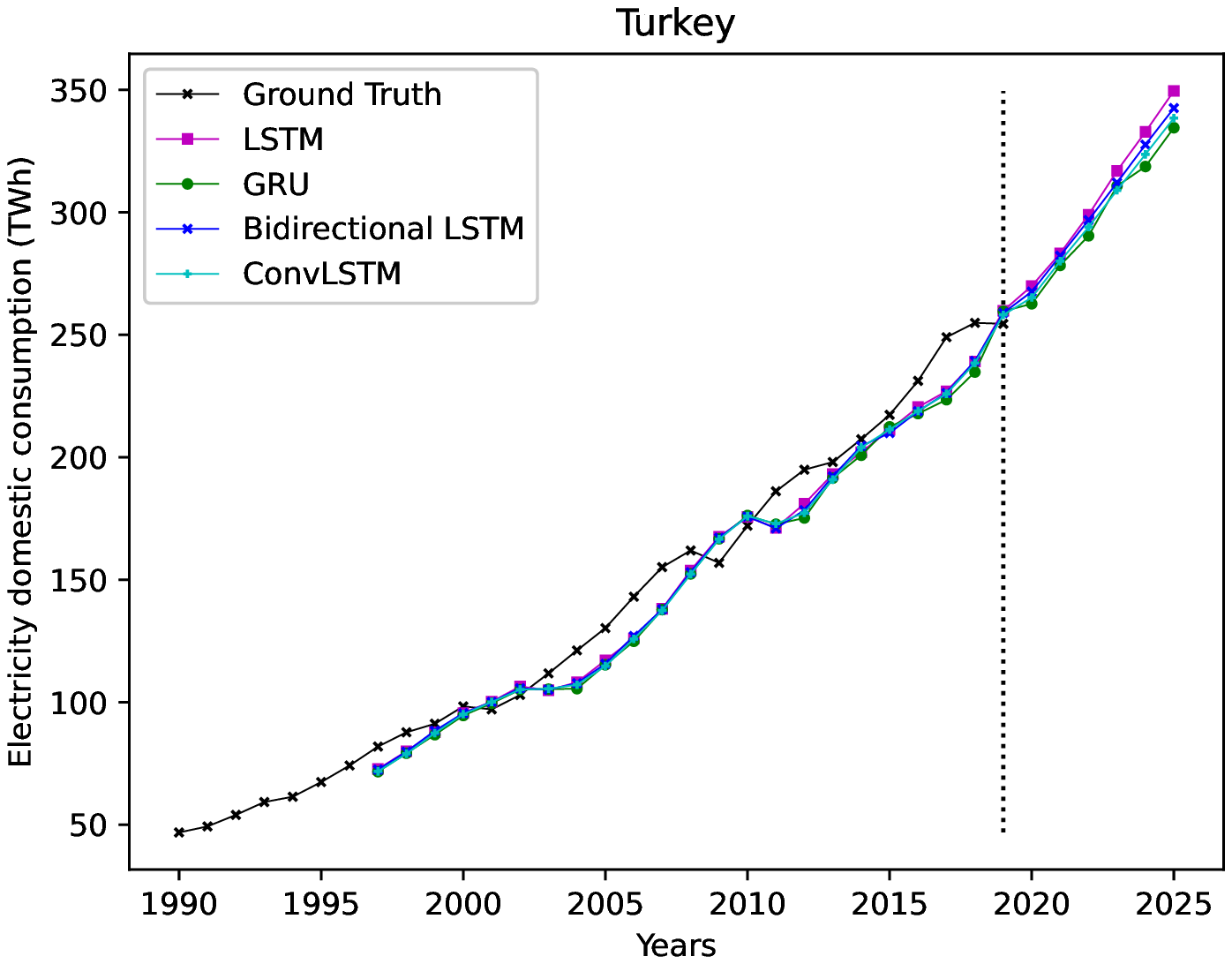}
\end{subfigure}
\end{figure}
\setcounter{figure}{8}
\begin{figure}[!h]
\begin{subfigure}[h]{\linewidth}
\hfil \includegraphics[width=5.2cm,height=4.2cm]{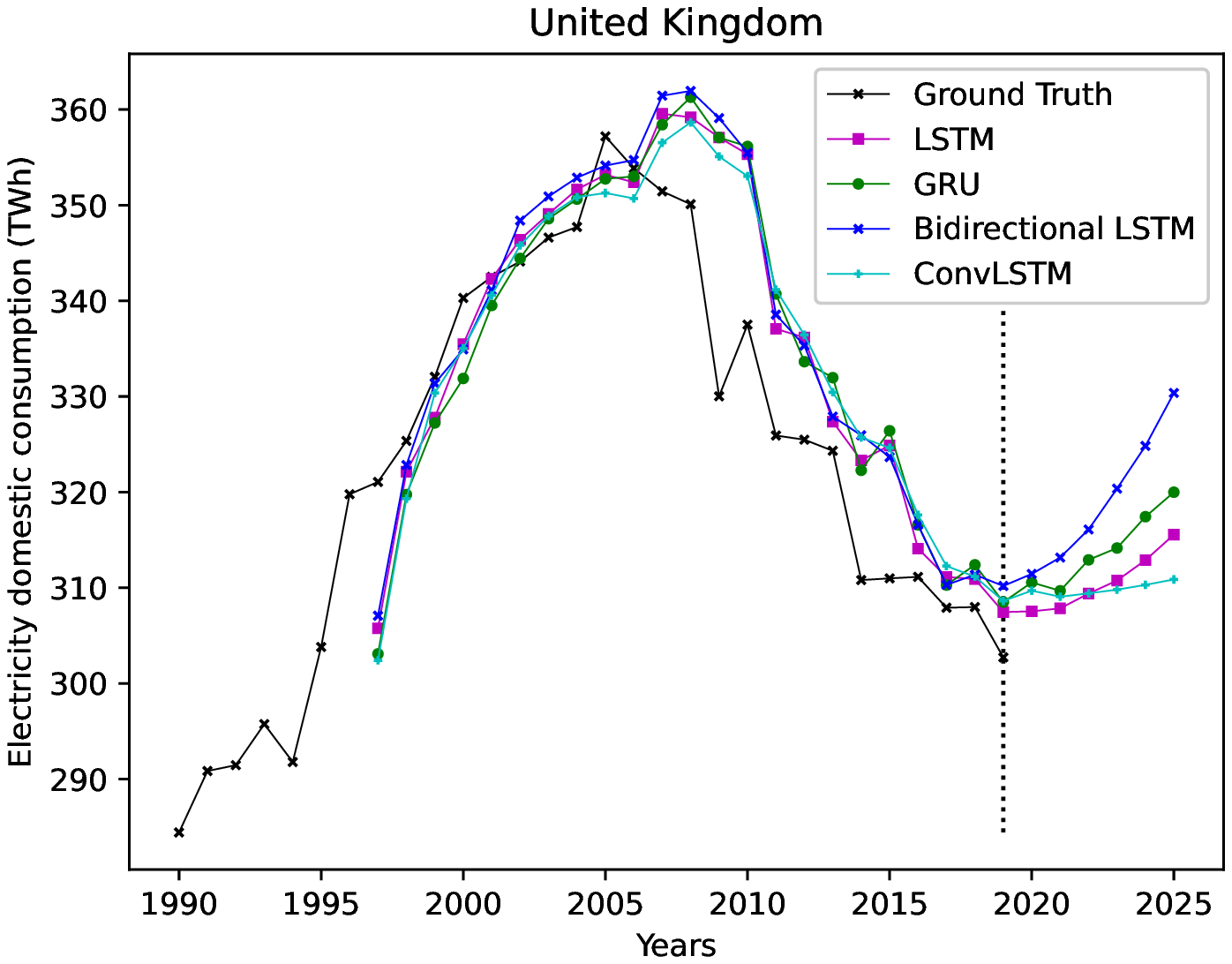}
\includegraphics[width=5.2cm,height=4.2cm]{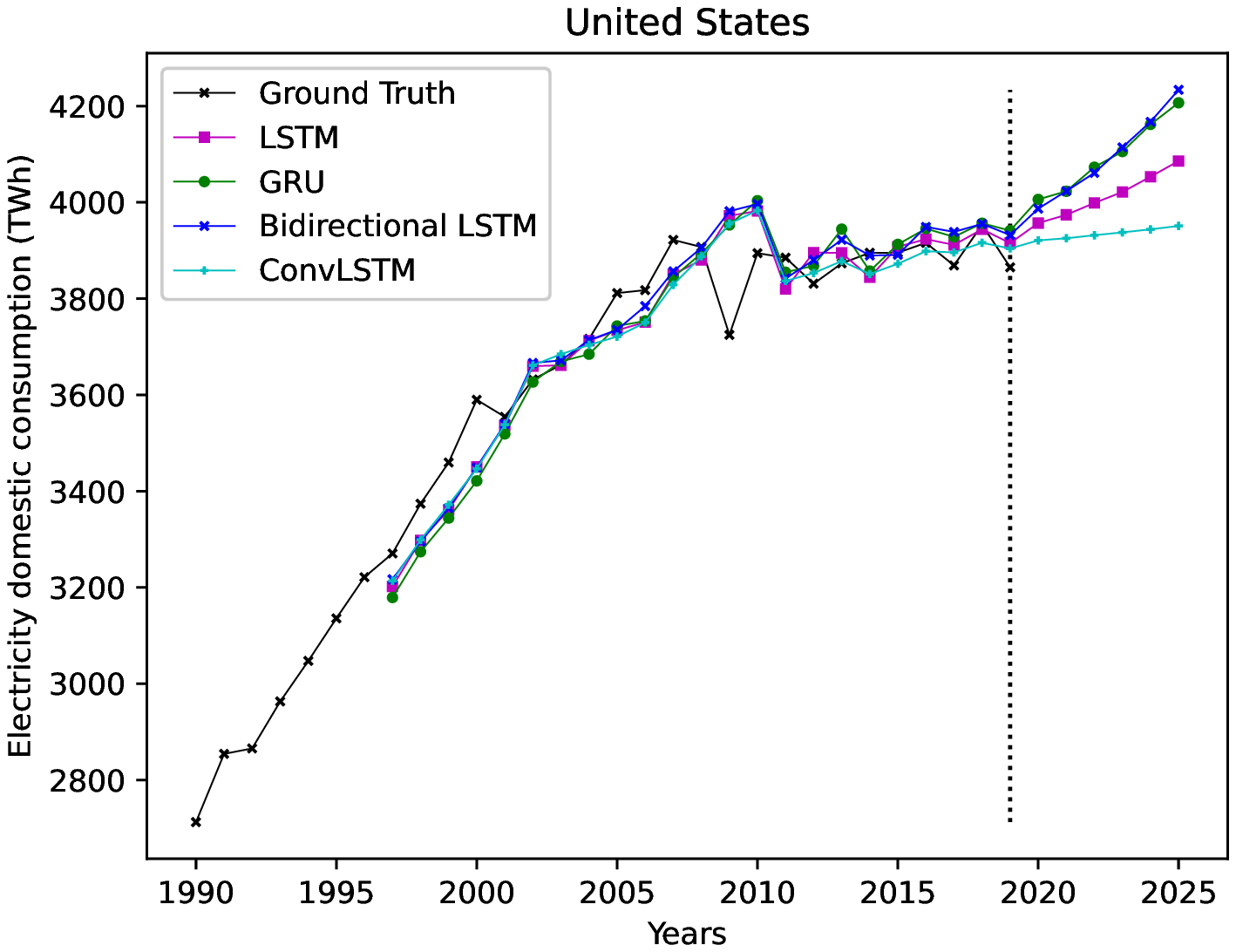}
\end{subfigure}
\caption{Predictions done by models trained with window size = 6\label{country}}
\end{figure}
\par  Seeing the test error rates, it can be concluded that we get the best performance for $window$ $size$ $= 6$ while using the LSTM based model. Ignoring the best model, all other models are not far away in terms of error rate. Overall, we can say that using $6$ as window gives us the optimum results compared to other window sizes. Fig. \ref{country} shows the actual predictions of the different models (trained with $window$ $size$ $= 6$) for all the G-20 members. Before the vertical line all model's performance can be compared to the ground truth at particular timestep (i.e. year). After the vertical line, up to the year $2025$ all models try to predict next values independently of each other. Years where previous true values are not available, model is self-fed it's own predicted values. Although this carries the error forward, we can get a rough estimate of next values models would actually predict if they were given true values.


\section{Conclusion}
In this paper we have estimated the 5 year electricity demand of G-20 Members utilizing the various RNN sets; such as- GRU, LSTM, Bidirectional LSTM and ConvLSTM. The best performance has been achieved using the LSTM based model by keeping the window size of 6. The result depicts, during the interval of (2019 – 2025); the energy demand of G – 20 Members can increase between the 118.69 to 155.8151 TWh. This result is rigorously based on the Terawatt hour of prediction utilising the Recurrent Neural Network Approach. During observations, the average energy growth from 2019 to 2025 of GRU, LSTM, Bidirectional LSTM, and Convolutional LSTM is achieved such as: 141.91, 140.36, 155.81 and 118.69 TWh respectively.  In future, these energy demand can be superseded using the various kind of Renewable Generating Sources such as – Solar Energy, Wind Energy, Hydro Energy and Tidal Energy for better fulfilment of supply – demand chain.

\section*{Acknowledgement}
We would like to thank Enerdata organization for allowing us to use the Energy Consumption Data prepared by them.

\bibliographystyle{unsrt}
\bibliography{main.bib}

\end{document}